\documentclass[letterpaper, 10 pt, conference]{ieeeconf}  

\IEEEoverridecommandlockouts                              

\usepackage{amsmath}
\usepackage{amssymb}
\usepackage{booktabs}
\usepackage{graphicx}
\usepackage{xcolor}
\usepackage{hyperref}

\title{\LARGE \bf
Large Reward Models: Generalizable Online Robot \\ Reward Generation with Vision-Language Models 
}

\author{Anonymous Authors
}
\author{
  Yanru Wu$^{1}$, Weiduo Yuan$^{1}$, Esteban Martinez Licon$^{1}$, Ang Qi$^{1}$, Vitor Guizilini$^{2}$, \\
  Jiageng Mao$^{1\dagger}$, and Yue Wang$^{1\dagger}$ \\[0.5em]
  $^{1}$ USC Physical Superintelligence Lab \quad $^{2}$ Toyota Research Institute \\
  $^{\dagger}$Equal advising \\
}

\begin{document}

\maketitle
\thispagestyle{empty}
\pagestyle{empty}

\begin{figure*}[t]
    \centering
    \includegraphics[width=1\linewidth]{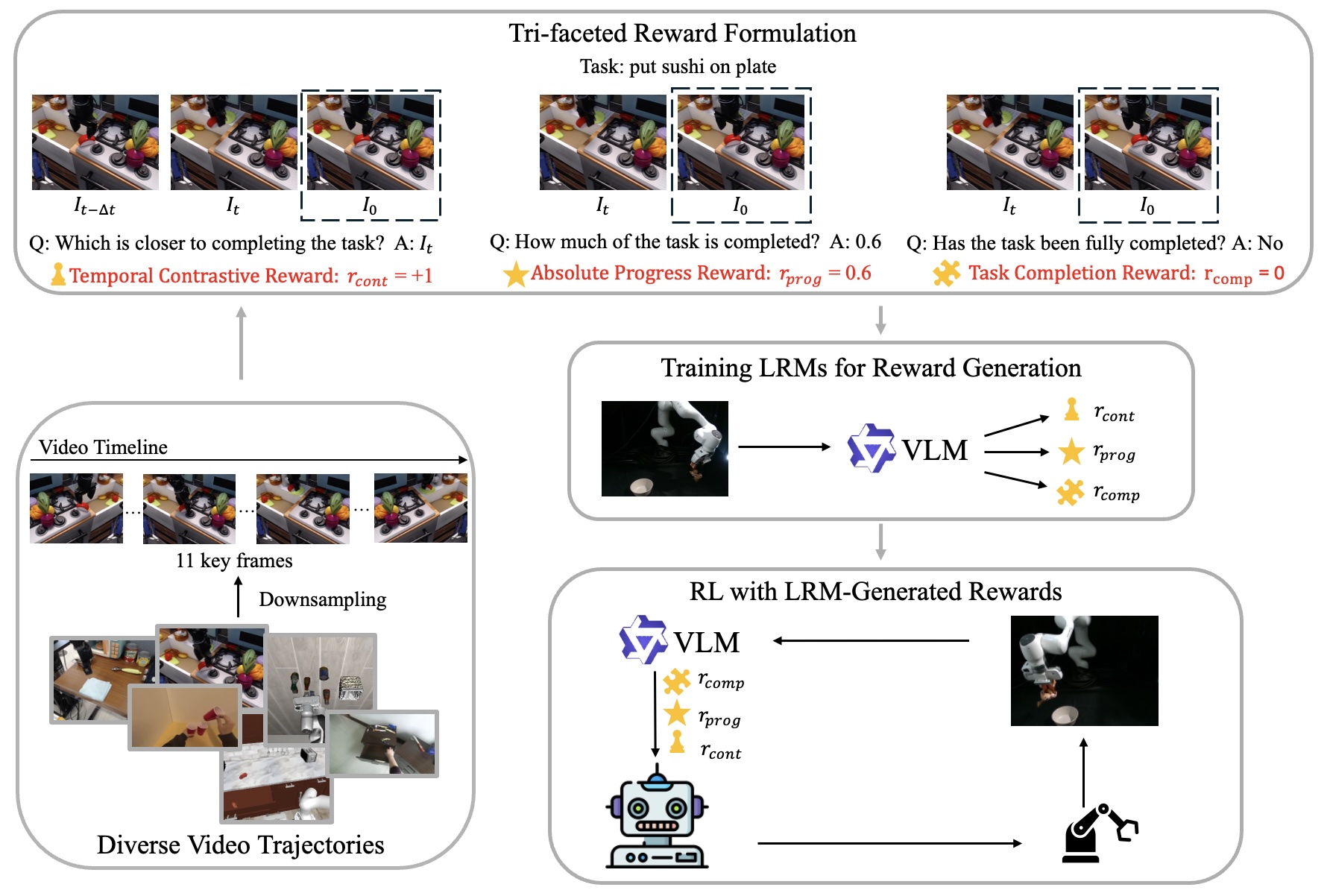}
    \caption{\textbf{Overview of Large Reward Models.} LRMs instantiate three modality-specific reward models from the same Qwen3-VL-8B-Instruct initialization: Temporal Contrastive Reward ($r_{cont}$) for relative temporal ranking, Absolute Progress Reward ($r_{prog}$) for dense progress estimation, and Task Completion Reward ($r_{comp}$) for terminal success verification. Completion and contrastive use separate LoRA adapters on a frozen backbone, while progress additionally updates the vision encoder and multimodal merger. The outputs represent separately trained reward interfaces rather
    than simultaneous predictions from a multi-task head. During single-interface deployment, one reward model is selected for the downstream setting, and its structured text response is parsed into scalar feedback.}
    \label{fig:overview}
\end{figure*}

\begin{abstract}

Reinforcement Learning (RL) has shown strong potential for improving robotic manipulation policies, yet its practical use remains bottlenecked by the difficulty of specifying reward functions that are both semantically meaningful and reusable across tasks. 
In this paper, we propose Large Reward Models (LRMs), a framework that adapts foundation VLMs into frame-level reward generators for robot policy refinement. 
We specialize a state-of-the-art VLM on a multi-source dataset spanning real-world robot trajectories, human-object interactions, and simulated manipulation environments. 
Unlike prior approaches that mainly evaluate trajectories post-hoc, LRMs expose multiple reward interfaces from visual observations: progress estimation, task completion, and temporal contrastive comparison. 
Starting from an imitation-learned policy, we use these VLM-derived rewards to guide PPO refinement on held-out long-horizon manipulation tasks. 
Our experiments show that LRM progress rewards provide the strongest non-privileged online refinement signal, improving the IL baseline and narrowing the gap to privileged environment rewards. 
We further deploy progress rewards for progress-weighted behavioral cloning on four real-world manipulation tasks spanning two robot platforms, improving over SFT on all four tasks. 
These results suggest that modality-specific specialization of foundation VLMs can provide practical visual reward signals for both simulated policy refinement and physical robot self-improvement without hand-coded task rewards. 

\end{abstract}

\section{INTRODUCTION}
The development of generalist robot policies has been accelerated by large-scale visuomotor and vision-language-action pre-training across diverse robotic datasets \cite{brohan2022rt,brohan2023rt,o2024open,team2024octo}. These policies provide strong imitation-learning (IL) backbones, but IL pre-training alone can still plateau on tasks requiring high precision, semantic judgment, or long-horizon coordination. Reinforcement learning (RL) offers a principled mechanism for post-training policy refinement and has shown progress in both real-world manipulation and broader robot learning pipelines \cite{levine2016end,zhu2020ingredients,mendonca2024continuously,luo2025precise,lei2025rl,intelligence2025pi}. Its effectiveness, however, remains tightly coupled to the availability of accurate, dense, and task-aware reward functions. In real-world and complex simulated environments, obtaining such rewards is still a major bottleneck, as traditional approaches often rely on labor-intensive human feedback or brittle task-specific reward engineering.

Learning visual rewards has therefore become a central direction for scalable robot policy improvement. Prior work has learned reward representations or temporal structure from demonstrations and human videos~\mbox{\cite{sermanet2018time,ma2022versatile,ma2022vip,ma2023liv}}, used language or vision-language models to generate reward programs, preferences, progress estimates, or token-probability rewards \cite{wang2024rl,ma2024eureka,ma2025vision,chen2026topreward,lee2026roboreward,liang2026robometer}, and adapted process-level or critic-style supervision for robotic manipulation \cite{lightman2024let,zhai2025vision,tan2025robo,ayalew2025progressor}. These methods demonstrate that foundation models can provide semantically meaningful feedback, but online policy refinement imposes additional interface constraints: rewards must be state-dependent, temporally informative, easy to parse into scalar feedback, and computationally practical during closed-loop interaction. This motivates directly specializing foundation VLMs into compact, frame-level reward interfaces rather than relying on a single monolithic reward format.

To bridge this gap, we introduce \textit{Large Reward Models (LRMs)}, a framework for adapting foundation VLMs into frame-level reward generators. We develop a suite of LRMs based on the Qwen3-VL \cite{bai2025qwen3} architecture and train them to formulate state-dependent rewards directly from visual observations. To support evaluation beyond the training domains, our models are trained on a balanced collection sampled from 24 data sources. This multi-domain dataset encompasses real-robot corpora (Open X-Embodiment \cite{o2024open}), human-object interaction data (HOI4D \cite{liu2022hoi4d}, EgoDex \cite{hoque2026egodex}), and simulated environments (LIBERO \cite{liu2023libero}, RoboCasa \cite{nasiriany2024robocasa}). 


LRMs provide \textit{Temporal Contrastive}, \textit{Absolute Progress}, and \textit{Task Completion} rewards for relative temporal ordering, dense progress shaping, and semantic success verification, respectively. This modular formulation allows us to compare reward modalities without assuming that their outputs should be directly combined.

We evaluate our LRM framework on held-out long-horizon manipulation benchmarks within ManiSkill3 \cite{tao2024maniskill3}. This benchmark serves as a testbed for coupling policy learning with learned visual rewards, allowing us to measure whether VLM-derived feedback can improve policies without privileged simulator rewards. We further deploy the progress interface for real-world weighted behavioral cloning on four physical manipulation tasks spanning two robot platforms.
In simulation, all three LRM reward interfaces achieve higher mean success rates than the IL baseline, with progress reward performing best; in the real world, LRM progress scoring improves over SFT on all four tasks.

In summary, our main contributions are as follows:
\begin{itemize}
\item We introduce \textit{Large Reward Models (LRMs)}, a framework that adapts foundation VLMs into frame-level visual reward generators for robot policy refinement.
\item The framework provides contrastive, progress, and completion interfaces that convert VLM outputs into parseable, state-dependent reward signals.
\item Experiments validate LRMs through online PPO refinement on held-out ManiSkill3 tasks and progress-weighted behavioral cloning across four real-world tasks on two robot platforms.
\item We analyze reward combination, query intervals, and inference latency for practical deployment.
\end{itemize}

\section{RELATED WORK}

\textbf{Policy refinement beyond imitation pre-training.}
Large-scale robot policies trained from diverse demonstrations and vision-language supervision have improved general-purpose manipulation \cite{brohan2022rt,brohan2023rt,o2024open,team2024octo}, yet precision-sensitive and long-horizon tasks still benefit from post-training refinement. Prior work explores simulation transfer, residual or interactive RL, real-world RL, and self-improving VLA pipelines \cite{tobin2017domain,peng2018sim,johannink2019residual,luo2024rlif,zhu2020ingredients,mendonca2024continuously,luo2025precise,lei2025rl,intelligence2025pi}. Across these approaches, constructing reusable visual reward signals remains an important challenge. LRMs complement these approaches by specializing VLMs as visual reward models.

\textbf{Learned visual reward models for robotics.}
To reduce manual reward engineering, prior methods learn rewards, representations, or temporal structure from demonstrations, human videos, and language-conditioned robot data \cite{sermanet2018time,ma2022versatile,ma2022vip,ma2023liv}. Recent foundation-model approaches further use LLMs and VLMs to generate reward code, infer preferences, estimate progress, or evaluate trajectory quality \cite{wang2024rl,ma2024eureka,ma2025vision,chen2026topreward,lee2026roboreward,liang2026robometer}. Concurrent work studies procedure-grounded progress, stage-aware value estimation, failure-aware dense rewards, transition-alignment rewards, and joint absolute-relative reward learning \cite{feng2026procvlm,chen2026sarm2,fang2026densereward,zhang2026dream2reward,chen2026vlm}. LRMs instead focus on modality-specific adaptation \cite{hu2021lora} of the same pretrained VLM into three separately deployable frame-level reward interfaces, evaluated through closed-loop PPO refinement and offline real-world adaptation.

\textbf{Process-level supervision and dense feedback.}
Sparse terminal rewards motivate intermediate progress signals. Process reward models provide step-level supervision \cite{lightman2024let}, while robotics methods use critic-style VLM feedback, process rewards, or self-supervised progress estimation \cite{zhai2025vision,tan2025robo,ayalew2025progressor}. LRMs build on this direction by training separate contrastive, progress, and completion interfaces from temporally grounded video supervision.

\section{METHODOLOGY}

While Reinforcement Learning (RL) can surpass imitation learning plateaus, its effectiveness depends on reward functions that provide both temporal resolution and semantic task understanding. To reduce the need for task-specific reward engineering, we propose Large Reward Models (LRMs), a framework that adapts foundation VLMs into frame-level visual reward generators.

Our framework utilizes Reinforcement Learning (RL) to refine a robotic policy $\pi_{\phi}$ by maximizing the expected return under a selected reward interface $m$:
\begin{equation}
J(\pi_{\phi}; m) = \mathbb{E}_{\tau \sim \pi_{\phi}} \left[ \sum_{t=0}^{T} \gamma^t r_m(\mathcal{I}_t^m, d) \right]
\label{eq:rl_objective}
\end{equation}
where $\mathcal{I}_t^m$ denotes the modality-specific visual input and $d$ is the task description. To obtain reward mappings without manual reward engineering, we construct a structured dataset $\mathcal{D} = \{(I_t, d, p_t)\}$ from unlabeled videos and specialize a foundation VLM into three reward interfaces: \textit{Temporal Contrastive} ($r_{cont}$), \textit{Absolute Progress} ($r_{prog}$), and \textit{Task Completion} ($r_{comp}$). Each interface maps visual observations and task language into parseable reward feedback, enabling LRMs to serve as frame-level reward providers for policy refinement.

\subsection{Extracting Reward Signals from Multi-domain Unlabeled Videos} 
To scale reward generation without manual reward labels, we extract supervision directly from a balanced subset of unlabeled video trajectories $V$ sampled from 24 sources. We use the temporal order of successful or demonstrative episodes as weak supervision: later frames are treated as higher-progress states than earlier frames. By mapping each sampled video timeline to a discrete progress scale, we transform raw frames into a structured dataset $\mathcal{D} = \{(I_t, d, p_t)\}$ for training visual reward models.

\textbf{Multi-domain Dataset Composition}: To improve transfer to held-out environments, we sample training episodes from 24 data sources covering diverse physical interactions, visual domains, and task semantics. We count sources at the dataset level rather than the collection-family level: each named Open X-Embodiment sub-dataset is counted independently because it corresponds to a distinct robot embodiment, workspace, collection protocol, or task distribution. The source pool contains 20 Open X-Embodiment real-robot sub-datasets \cite{o2024open}, 2 egocentric human-object interaction datasets (HOI4D \cite{liu2022hoi4d} and EgoDex \cite{hoque2026egodex}), and 2 simulated benchmarks (LIBERO \cite{liu2023libero} and RoboCasa \cite{nasiriany2024robocasa}).

\textbf{Temporal Progression Sampling}: To construct supervision for the three reward interfaces, we sample each episode at normalized times $p \in \{0.0, 0.1, 0.2, \dots, 0.9, 1.0\}$, yielding 11 keyframes spanning the recorded execution. Each frame is paired with its temporal label and the episode's original task description $d$. Normalizing the timeline places episodes of different durations on a common scale and provides supervision without manual frame-level reward annotation. 

\subsection{Tri-faceted Reward Formulation}
To bridge high-level semantic instructions and the feedback requirements of policy optimization, we formulate three reward interfaces from the temporal progress of multi-domain videos. The interfaces are complementary but not assumed to be naturally commensurate: each is intended to support a different use case in robot policy refinement.

The \textit{Temporal Contrastive Reward} is designed to provide a relative directional gradient by comparing the current state against a previous observation. Given the current frame $I_t$, a previous frame $I_{t-\Delta t}$, and the task description $d$, this format evaluates which state is semantically closer to the task goal. The qualitative progression is mapped to a discrete numerical reward $r_{cont}$:\begin{equation}
r_{cont}=
\begin{cases}
    +1.0, & \text{if } I_t \text{ is closer to the goal than } I_{t-\Delta t} \\
    -1.0, & \text{if } I_{t-\Delta t} \text{ is closer to the goal than } I_t \\
     0.0, & \text{if no valid selection is produced}
\end{cases}
\label{eq:contrastive_cases}
\end{equation}
By gauging relative progress between frames, this formulation avoids the need for globally calibrated absolute scores and provides directional feedback about local task improvement.

To provide spatial-temporal grounding, the \textit{Absolute Progress Reward} is formulated as ordinal progress prediction over discretized normalized progress bins. Given the current observation $I_t$, optionally with the initial state $I_0$ as a visual anchor, and the task description $d$, the reward signal $r_{prog}$ represents a normalized completion score:
\begin{equation}
r_{prog} \in \{0.0, 0.1, 0.2, \dots, 0.9, 1.0\}
\end{equation}
The eleven bins represent normalized temporal progress under the training supervision, allowing the interface to assign graded scores to intermediate observations. 

The \textit{Task Completion Reward} provides a definitive binary assessment of the current state $I_t$. Given the observation $I_t$, optionally with the initial state $I_0$ as visual context, and task description $d$, the signal is defined as a binary classification to determine if the semantic requirements of the goal have been fully satisfied:\begin{equation}
r_{comp} =
\begin{cases}
    1, & \text{if semantic requirements are met} \\
    0, & \text{otherwise}
\end{cases}
\label{eq:completion_cases}
\end{equation}
By outputting a binary success signal $r_{comp}$, this modality provides a semantic terminal judgment that is useful for success verification, sparse reward assignment, and offline trajectory filtering.

\subsection{Training LRMs for Reward Generation}
We initialize each LRM from Qwen3-VL-8B-Instruct and train the three reward modalities separately on the structured dataset $\mathcal{D}$. Completion and contrastive update distinct Low-Rank Adaptation (LoRA) adapters \cite{hu2021lora} while keeping the pretrained backbone frozen. Progress updates its LoRA adapter together with the vision encoder and multimodal merger, while the remaining backbone parameters stay frozen. We do not introduce new reward tokens or architectural reward heads. Each model generates text using the normal autoregressive vocabulary, and a deterministic parser converts the structured output into a scalar or binary reward. Prompting specifies the visual context, task instruction, and output schema rather than task-specific hand-coded reward rules. For all three modalities, training mixes anchored and unanchored visual inputs: unanchored inputs contain only the minimal observation set for that modality, while anchored inputs additionally include the episode's initial frame $I_0$ as visual context. Thus, progress and completion examples may use either $I_t$ or $(I_0, I_t)$, while contrastive examples may use either $(I_{t-\Delta t}, I_t)$ or $(I_0, I_{t-\Delta t}, I_t)$. Contrastive deployment additionally generates a short physical comparison before the final reward. The zero-shot Qwen3-VL baselines use the original pretrained weights with the same prompts and parsers.

The \textit{contrastive discrimination model} is trained as a preference judge to provide relative directional gradients by comparing the visual input $\mathcal{I}^{cont}$, which contains the two temporal frames $I_{t-\Delta t}$ and $I_t$ in randomized order and may additionally include the initial frame $I_0$. We utilize Direct Preference Optimization (DPO) \cite{rafailov2023direct} to align the model's preferences with the temporal progression recorded in $\mathcal{D}$ using the following objective:
\begin{equation}
\begin{split}
L_{\text{DPO}}(\theta; \pi_{\text{ref}}) = -\mathbb{E} \Big[ \log \sigma \Big( & \beta \log \frac{\pi_\theta(y_w \mid \mathcal{I}^{cont}, d)}{\pi_{\text{ref}}(y_w \mid \mathcal{I}^{cont}, d)} \\
& - \beta \log \frac{\pi_\theta(y_l \mid \mathcal{I}^{cont}, d)}{\pi_{\text{ref}}(y_l \mid \mathcal{I}^{cont}, d)} \Big) \Big]
\end{split}
\label{eq:dpo_loss}
\end{equation}
where $(y_w, y_l)$ are the preferred and rejected text responses induced by temporal order. We ask the model to produce a short physical comparison before outputting $r_{cont} \in \{+1, -1, 0\}$, so the reward is based on task-relevant object interactions rather than low-level visual changes.

The \textit{progress estimation} and \textit{completion judgment} models are both trained via Supervised Fine-Tuning (SFT) to map their visual inputs to $r_{prog} \in \{0.0, \dots, 1.0\}$ and $r_{comp} \in \{0, 1\}$, respectively. For the \textit{progress estimation model}, we maximize the likelihood of a joint reasoning-and-label sequence:
\begin{equation}
\mathcal{L}_{\text{prog}}
=
-\mathbb{E}\left[
\log P(\mathrm{CoT}, r_{\text{prog}}
\mid \mathcal{I}^{prog}, d)
\right]
\label{eq:prog_loss}
\end{equation}
This reasoning-first supervision encourages the model to articulate physical cues before concluding $r_{prog}$. Although CoT is used for supervision, progress inference returns only the final structured score. In contrast, the \textit{completion judgment model} is optimized to directly predict the binary success signal:
\begin{equation}
\mathcal{L}_{\text{comp}} = -\mathbb{E} \left[ \log P(r_{\text{comp}} \mid \mathcal{I}^{comp}, d) \right]
\label{eq:comp_loss}
\end{equation}
This provides a terminal signal for goal verification and policy refinement.

\subsection{Online Policy Refinement via LRM-Generated Rewards}
To move beyond imitation learning plateaus, we use the frozen LRMs as visual reward providers for refining a base policy $\pi_{\phi}$ initialized from $\pi_{SFT}$. This stage uses closed-loop interaction and PPO updates, while keeping the reward model fixed.

\textit{Online Interaction and Reward Integration}: During each episode, the policy interacts with the environment by sampling actions $a_t \sim \pi_{\phi}(a_t \mid I_t, d)$ based on the current visual observation $I_t$ and task description $d$. To bridge the computational gap between VLM inference and control, we use an Interval-Hold strategy. The selected LRM is queried every $K$ environment steps to compute $r_m = \text{LRM}_m(\mathcal{I}_t^m, d)$ for one reward modality $m \in \{cont, prog, comp\}$. 
\begin{equation}
r_t = w_{m}r_{m}, 
\end{equation}
where $w_m$ is the scaling factor for the selected modality. The reward $r_t$ is cached and held for $K$ steps. Each single-interface policy refinement run uses one selected reward model and its modality-specific parameters.

\textit{Policy Optimization and Refinement}: We utilize Proximal Policy Optimization (PPO) to update the policy parameters $\phi$. The policy is optimized with the clipped surrogate objective:

\begin{equation}
\mathcal{L}_{\mathrm{PPO}}(\phi)
= -\mathbb{E}_{t}\!\left[
\min\!\left(\rho_t\hat{A}_t,\bar{\rho}_t\hat{A}_t\right)
\right],
\label{eq:ppo_loss}
\end{equation}
where
$\rho_t=\pi_{\phi}(a_t\mid I_t,d)/
\pi_{\phi_{\mathrm{old}}}(a_t\mid I_t,d)$
is the probability ratio between the updated and behavior policies,
and $\bar{\rho}_t=\operatorname{clip}(\rho_t,1-\epsilon,1+\epsilon)$
is its clipped value. To ensure stable convergence and handle the dense nature of LRM feedback, we employ Generalized Advantage Estimation (GAE) to compute the advantage $\hat{A}_t$:
\begin{equation}
\hat{A}_t = \sum_{l=0}^{T-t-1} (\gamma \lambda)^l \delta_{t+l}, \quad \delta_t = r_t + \gamma V_{\psi}(I_{t+1}) - V_{\psi}(I_t),
\label{eq:gae_advantage}
\end{equation}
where $\delta_t$ is the TD error computed from the LRM reward $r_t$,
enabling optimization from visual feedback rather than privileged
simulator state or hand-coded rewards.

\section{EXPERIMENTAL RESULTS}
We evaluate LRMs at both the reward-model and policy levels. We first compare each specialized reward interface with zero-shot Qwen3-VL using identical prompts and parsers. We then assess downstream utility through closed-loop PPO refinement on held-out ManiSkill3 \cite{tao2024maniskill3} tasks and analyze reward alignment with privileged simulator signals on collected rollouts. Query-interval and latency experiments quantify the computational trade-offs of online deployment. Finally, we evaluate progress-weighted behavioral cloning on four real-world tasks across two robot platforms, demonstrating an offline use of LRM feedback for policy adaptation.
\subsection{Evaluating Large Reward Models (LRMs)}
Before policy refinement, we evaluate whether specialization improves reward prediction.
We test the three reward modalities—$r_{cont}$, $r_{prog}$, and $r_{comp}$—against the zero-shot Qwen3-VL-8B-Instruct baseline on held-out evaluation sets. This isolates modality-specific specialization from prompt engineering because the baseline uses the original pretrained weights with the same reward prompts and parsers.

\textit{Contrastive Discrimination Model}:
Table~\ref{tab:contrastive_results} compares the zero-shot baseline with our fine-tuned LRM on the generation quality of the \textit{Temporal Contrastive Reward}, $r_{cont} = \mathrm{LRM}_{cont}(\mathcal{I}^{cont}, d)$. 
The ranking correlation metrics improve: both Kendall's $\tau$ and Spearman's $\rho$ increase by 15.3\%. These metrics capture the model's ability to produce consistent temporal orderings required for a stable directional gradient.

\begin{table}[h]
\centering
\setlength{\tabcolsep}{3pt}
\caption{\textbf{Contrastive Discrimination performance.} LRM specialization improves ranking correlation as measured by Kendall's $\tau$ and Spearman's $\rho$, indicating more consistent temporal ordering.}
\label{tab:contrastive_results}
\begin{tabular}{lccc}
\toprule
Model  & Kendall's $\tau$ & Spearman's $\rho$ \\
\midrule
Qwen3-VL & 0.257 & 0.257 \\
Our LRM  & \textbf{0.296} & \textbf{0.296} \\
\midrule
\textit{Improvement}  & \textbf{+15.3\%} & \textbf{+15.3\%} \\
\bottomrule
\end{tabular}
\end{table}

\begin{figure}
    \centering
    \includegraphics[width=1\linewidth]{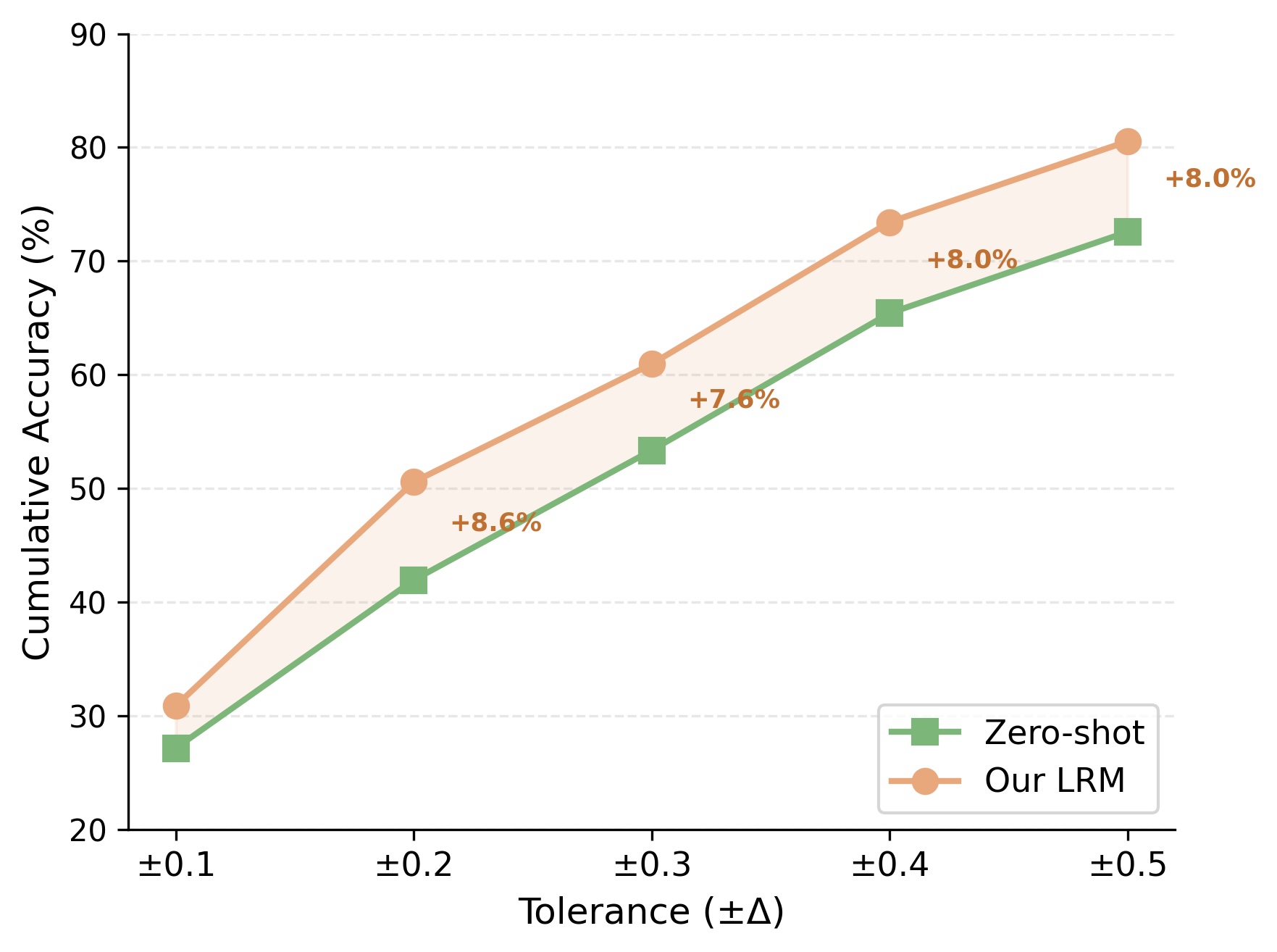}
    \caption{\textbf{Cumulative accuracy at varying tolerance thresholds ($\pm\Delta$).} LRM improves over zero-shot Qwen3-VL by 8.63 percentage points at $\pm0.2$.}
    \label{fig:cumulative_accuracy}
\end{figure}

\textit{Progress Estimation Model}:
Table~\ref{tab:progress_estimation} reports progress prediction against temporal proxy labels. Relative to zero-shot Qwen3-VL, LRM reduces MAE from 0.378 to 0.302 and RMSE from 0.490 to 0.395, while Acc@$\pm0.2$ increases from 41.95\% to 50.58\%. Exact accuracy changes more modestly, from 13.10\% to 13.87\%, so the gain is more apparent when allowing a tolerance around the target label. Figure~\ref{fig:cumulative_accuracy} presents this comparison across tolerance thresholds. 
Together, these metrics show that LRM specialization improves agreement with the temporal progress labels relative to the base VLM.

\begin{table}[h]
\centering
\caption{\textbf{Progress estimation against temporal proxy labels.} LRM has lower MAE and RMSE and higher prediction accuracy than zero-shot Qwen3-VL.}
\label{tab:progress_estimation}
\begin{tabular}{lccc}
\toprule
\textit{Metric} & Qwen3-VL & Our LRM & $\Delta$ \\
\midrule
Exact Acc    & 13.10\% & \textbf{13.87\%} & +0.77\% \\
Acc@$\pm$0.2  & 41.95\% & \textbf{50.58\%} & +8.63\% \\
MAE        & 0.378   & \textbf{0.302}   & \textbf{$-$20.0\%} \\
RMSE       & 0.490   & \textbf{0.395}   & \textbf{$-$19.3\%} \\
\bottomrule
\end{tabular}
\end{table}

\textit{Task Completion Model}:
The task completion model achieves 69.38\% accuracy, compared to 69.23\% for zero-shot Qwen3-VL. Specialization provides little additional benefit on this evaluation set, consistent with a strong base-model prior for goal recognition.

\subsection{Policy Refinement on the ManiSkill3 Benchmark}
Starting from a $\pi_{0.5}$ SFT baseline, we evaluate policy refinement on ManiSkill3~\cite{tao2024maniskill3} for 30 RL iterations ($\approx$768,000 environment steps). We organize the closed-loop results into two comparisons. Table~\ref{tab:closed_loop_main} compares our reward interfaces against zero-shot Qwen3-VL reward prompts, a privileged environment reward upper bound, and a tri-reward combination. Table~\ref{tab:progress_reward_compare} separately compares LRM-$r_{prog}$ with prior progress-style VLM reward models.

\textbf{Closed-Loop Performance Evaluation}: Table~\ref{tab:closed_loop_main} shows that the strongest LRM reward is LRM-$r_{prog}$, which reaches $58.94\% \pm 2.18\%$ from a $55.62\% \pm 3.30\%$ SFT baseline. This improves over the IL baseline while still remaining below the privileged environment reward upper bound ($60.69\% \pm 1.90\%$). The benefit of specialization is clearest for progress and contrastive rewards: LRM-$r_{prog}$ improves over the corresponding zero-shot Qwen3-VL progress prompt by 4.57 percentage points, while LRM-$r_{cont}$ improves over the zero-shot contrastive prompt by 3.25 percentage points. In contrast, the base and adapted completion models achieve similar mean performance, consistent with the intrinsic completion evaluation above.

\begin{table}[h]
\centering
\caption{\textbf{Closed-loop success rate (\%) on ManiSkill3.} Mean $\pm$ standard deviation (SD) over 5 evaluation seeds for 1 trained checkpoint per condition. LRM-$r_{prog}$ has the highest non-privileged mean success rate.}
\label{tab:closed_loop_main}
\begin{tabular}{llc}
\toprule
Group & Method & Success Rate \\
\midrule
\multicolumn{2}{l}{$\pi_{0.5}$ SFT Baseline} & $55.62 \pm 3.30$ \\
\midrule
Privileged & Env Reward & $60.69 \pm 1.90$ \\
\midrule
Zero-shot VLM & Qwen3-VL-$r_{comp}$ & $57.63 \pm 2.53$ \\
Zero-shot VLM & Qwen3-VL-$r_{cont}$ & $53.31 \pm 1.20$ \\
Zero-shot VLM & Qwen3-VL-$r_{prog}$ & $54.37 \pm 2.53$ \\
\midrule
LRM & LRM-$r_{comp}$ & $57.00 \pm 2.22$ \\
LRM & LRM-$r_{cont}$ & $56.56 \pm 1.15$ \\
LRM & LRM-$r_{prog}$ & $\mathbf{58.94 \pm 2.18}$ \\
\midrule
Combined & Tri-reward LRM & $55.31 \pm 1.94$ \\
\bottomrule
\end{tabular}
\end{table}

The Tri-reward LRM directly sums the three reward outputs and reaches $55.31\% \pm 1.94\%$, below all individual LRM interfaces. These results favor selecting an interface for the downstream setting over simply mixing all three outputs: progress supplies intermediate feedback for policy improvement, completion supports terminal success verification, and contrastive comparison supplies relative temporal ordering.

Table~\ref{tab:progress_reward_compare} compares prior progress reward models under the same PPO protocol and query interval ($K=10$). LRM-$r_{prog}$ achieves the highest mean success rate in this setting.

\textbf{Open-Loop Reward Quality Analysis}: We assess reward alignment on 80 trajectories using privileged simulator signals. Each LRM remains fixed; the SFT and RL columns identify the policies generating the evaluated trajectories. For absolute rewards, we report ROC-AUC, pairwise accuracy, and both global and per-trajectory Pearson correlation. These measures characterize discriminative performance and temporal agreement at complementary levels. For contrastive reward, step-level metrics assess the direction of local changes and the monotonicity of accumulated feedback.

In Table~\ref{tab:openloop_comp_prog}, completion and progress ROC-AUC are higher on RL-policy rollouts (0.795 and 0.950) than on SFT-policy rollouts (0.660 and 0.874). Pairwise accuracy follows the same pattern, increasing from 45.4\% to 63.9\% for completion and from 80.1\% to 93.4\% for progress. Progress reaches a per-trajectory Pearson correlation of 0.671 on RL rollouts, indicating positive agreement within individual executions as well as in the pooled observations. Completion also exhibits positive alignment, with global and per-trajectory correlations of 0.757 and 0.331, respectively.

\begin{table}[h]
\centering
\caption{\textbf{Progress reward comparison on ManiSkill3 (\%).} All methods use the same PPO setting and $K=10$. Mean $\pm$ SD over 5 evaluation seeds for 1 trained checkpoint per method; LRM-$r_{prog}$ has the highest mean success rate.}
\label{tab:progress_reward_compare}
\begin{tabular}{lc}
\toprule
Progress Reward Model & Success Rate \\
\midrule
RoboReward-8B \cite{lee2026roboreward} & $53.37 \pm 1.85$ \\
Robometer-4B \cite{liang2026robometer} & $57.37 \pm 1.18$ \\
TOPReward \cite{chen2026topreward} & $51.06 \pm 1.63$ \\
\midrule
LRM-$r_{prog}$ & $\mathbf{58.94 \pm 2.18}$ \\
\bottomrule
\end{tabular}
\end{table}

\begin{table}[h]
\centering
\caption{\textbf{Open-loop reward alignment.} Fixed LRMs show higher agreement with simulator signals on RL-policy rollouts than on SFT-policy rollouts. Columns identify the rollout policy for each reward interface.}
\label{tab:openloop_comp_prog}
\begin{tabular}{lcccc}
\toprule
 & \multicolumn{2}{c}{$r_{comp}$} & \multicolumn{2}{c}{$r_{prog}$} \\
\cmidrule(lr){2-3}\cmidrule(lr){4-5}
Metric & SFT & RL & SFT & RL \\
\midrule
ROC-AUC & 0.660 & \textbf{0.795}  & 0.874 & \textbf{0.950}\\
Pairwise Acc (\%) & 45.4 & \textbf{63.9}  & 80.1 & \textbf{93.4}\\
Global Pearson & 0.398 & \textbf{0.757}  & 0.739 & \textbf{0.773}\\
Per-traj Pearson & 0.257 & \textbf{0.331}  & 0.577 & \textbf{0.671}\\
\bottomrule
\end{tabular}
\end{table}

The fixed contrastive LRM maintains similar directional accuracy across the two rollout sets (85.2\% and 85.5\%; Table~\ref{tab:openloop_cont}). Progress recall changes from 85.6\% to 86.0\%, while cumulative-reward monotonicity on successful trajectories increases from 0.535 to 0.559. Because the reward model remains fixed, these differences reflect the rollout distribution after refinement, in which the policy visits high-progress and successful states more frequently. Together with the absolute-reward results, these measurements indicate that the three LRM interfaces provide task-relevant feedback that is broadly consistent with privileged simulator signals across both SFT and refined-policy rollouts.

\begin{table}[h]
\centering
\caption{\textbf{Open-loop step-level reliability ($r_{cont}$).} The fixed contrastive LRM shows similar directional accuracy on SFT and RL rollouts, with slightly higher metrics on RL rollouts.}
\label{tab:openloop_cont}
\begin{tabular}{lcc}
\toprule
Metric & SFT &  RL-$r_{cont}$ \\
\midrule
Direction Acc (\%) & 85.2 & \textbf{85.5} \\
Progress Recall (\%) & 85.6 & \textbf{86.0} \\
Monotonicity (success) & 0.535 & \textbf{0.559} \\
\bottomrule
\end{tabular}
\end{table}

\subsection{Query Interval and Inference Cost}
Table~\ref{tab:k_sweep} evaluates the effect of the reward query interval $K$. Decreasing $K$ from 10 to 5 doubles the query frequency but does not consistently improve success rates. Completion and progress remain relatively stable across the tested intervals and achieve their highest mean performance at $K=10$, whereas contrastive reward has a lower mean at $K=16$. This pattern is consistent with frequent queries observing nearly identical states, while long holds can apply earlier pairwise judgments to states that have already changed. We therefore use $K=10$, which halves the query frequency relative to $K=5$ while maintaining performance and avoiding the degradation observed for contrastive reward at $K=16$.

\begin{table}[h]
\centering
\caption{\textbf{Query interval sweep (\%).} Mean $\pm$ SD over 5 evaluation seeds for 1 trained checkpoint per configuration. Progress and completion remain stable across the tested intervals, while contrastive reward drops at $K=16$, supporting $K=10$ as a practical trade-off.}
\label{tab:k_sweep}
\resizebox{\columnwidth}{!}{
\begin{tabular}{lccc}
\toprule
Reward & $K=5$ & $K=10$ & $K=16$ \\
\midrule
$r_{comp}$ & $55.50 \pm 2.83$ & $\mathbf{57.00 \pm 2.22}$ & $56.00 \pm 0.81$ \\
$r_{cont}$ & $\mathbf{57.12 \pm 2.05}$ & $56.56 \pm 1.15$ & $52.62 \pm 1.42$ \\
$r_{prog}$ & $57.81 \pm 3.68$ & $\mathbf{58.94 \pm 2.18}$ & $56.81 \pm 2.65$ \\
\bottomrule
\end{tabular}
}
\end{table}

Table~\ref{tab:latency} reports single-query HTTP round-trip latency on an NVIDIA RTX 6000 Ada GPU using representative rollout frames, excluding model loading and server startup. Progress and completion run in the 340--410 ms range with little LoRA overhead, whereas contrastive reward is much slower because it compares two observations and generates the only physical reasoning response. These measurements favor progress or completion for frequent online reward queries and reserve contrastive comparison for settings where relative ordering is worth the additional inference cost.

\begin{table}[h]
\centering
\caption{\textbf{Single-query reward inference latency.} HTTP round-trip time in ms (mean $\pm$ SD). Progress and completion are sub-second; contrastive queries include reasoning generation and are substantially slower.}
\label{tab:latency}
\begin{tabular}{lcc}
\toprule
Reward Query & Qwen3-VL & LRM \\
\midrule
Progress & $409 \pm 9$ & $360 \pm 19$ \\
Completion & $342 \pm 6$ & $340 \pm 5$ \\
Contrastive & $6838 \pm 2184$ & $7093 \pm 2307$ \\
\bottomrule
\end{tabular}
\end{table}

\begin{figure*}
    \centering
    \includegraphics[width=1\linewidth]{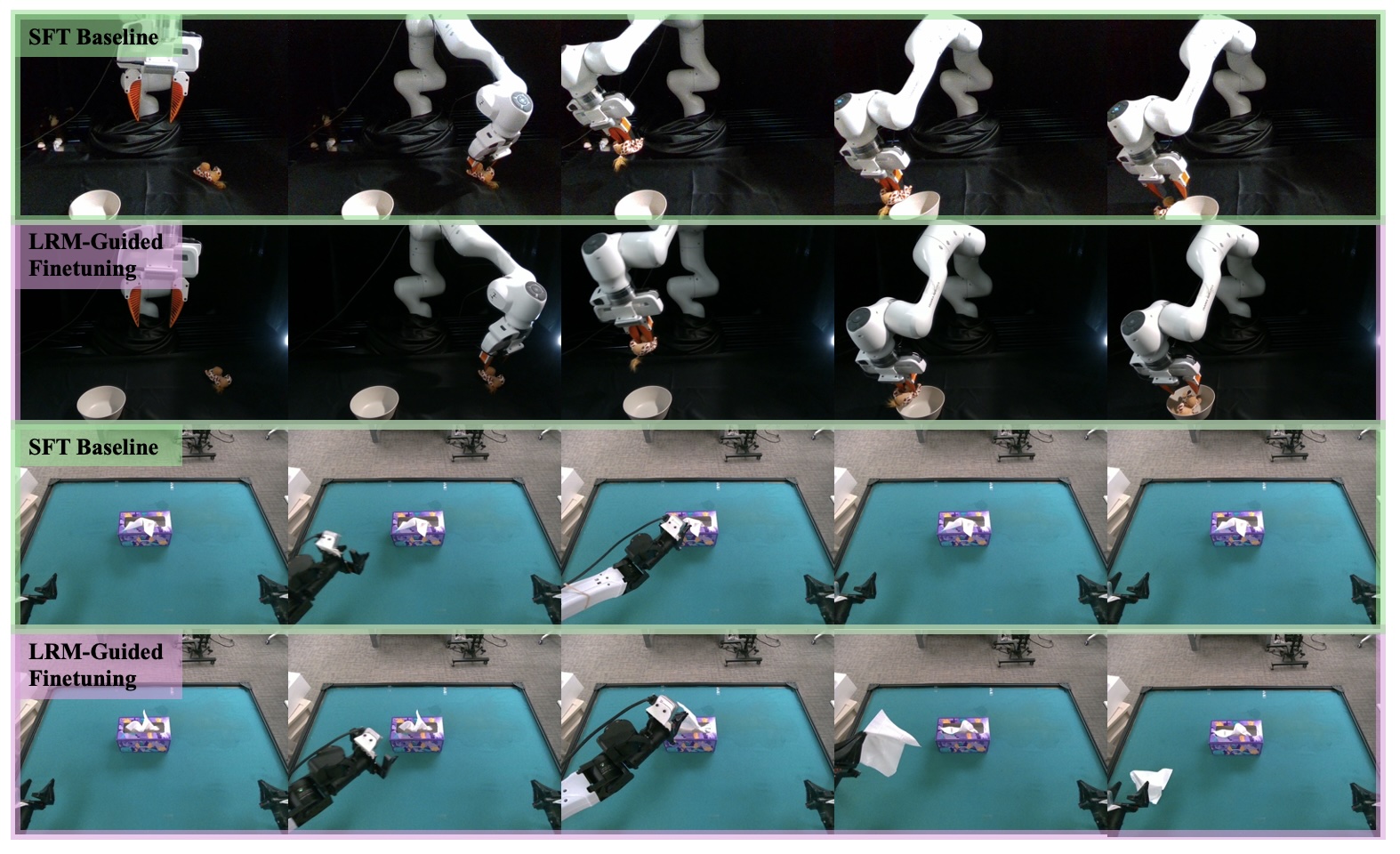}
    \caption{\textbf{Qualitative comparison of real-world rollouts.} On the giraffe-in-bowl task, the SFT baseline places the toy beside the target bowl, whereas the policy after LRM-guided progress-weighted fine-tuning places it inside the bowl. On the tissue-out-of-box task, the SFT baseline fails to grasp the tissue, whereas the policy after LRM-guided progress-weighted fine-tuning successfully grasps and pulls it out of the box. These examples illustrate that LRM-guided adaptation corrects distinct policy failure modes and improves real-world task execution.}
    \label{fig:rollout_compare}
\end{figure*}

\subsection{Real-World Policy Self-Improvement}
We further test whether progress rewards can improve physical robot policies through offline adaptation. Instead of querying a VLM reward online during robot control, we use progress scores to weight collected rollouts for behavioral cloning (BC). 

We evaluate four tasks across two robot platforms: \emph{pick up the toy giraffe and place it in the bowl}, \emph{stack the can on the cube}, \emph{open a drawer}, and \emph{pull a tissue out of a tissue box}. Starting from the SFT policy, we collect 60 real-world rollouts per task. We score the same rollouts separately with zero-shot Qwen3-VL and LRM progress, using identical prompts and parsers. For each rollout, we sample 5 frames uniformly across the full trajectory and 5 frames from the final 60 frames, remove duplicates, and score the sampled observations with $r_{prog}$. Let $P_e$ denote the sampled progress scores for episode $e$; we define the episode score as $s_e=\mathrm{mean}(\mathrm{top3}(P_e))$. This top-3 mean is less sensitive to isolated score noise than the peak score and less sensitive to terminal occlusion than the final sampled score.

We oversample high-confidence rollouts as full trajectories, retain high-progress rollouts as prefixes ending shortly after peak progress, and discard low-progress rollouts. Trajectory duplication implements weighted sampling with the standard BC objective.

Table~\ref{tab:robot_success_rate} shows that LRM-scored BC consistently improves over the SFT baseline across all four tasks, with gains ranging from 11.7 to 80.0 percentage points. In contrast, zero-shot Qwen3-VL scoring produces task-dependent outcomes: it improves performance on giraffe-in-bowl and tissue-out-of-box, but reduces success by 11.6 and 28.4 percentage points on can-on-cube and open-drawer. Although Qwen3-VL achieves the highest result on tissue-out-of-box, its mixed effects indicate that LRM specialization provides a more consistent direction of policy improvement across manipulation tasks and robot platforms. Figure~\ref{fig:rollout_compare} presents representative qualitative examples.

\begin{table}[h]
\centering
\caption{\textbf{Real-world progress-weighted BC.} Success rates (\%) with successful trials out of 60. Across two robot platforms, LRM scoring consistently improves over SFT on all four tasks, whereas zero-shot Qwen3-VL scoring produces task-dependent improvements and degradations.}
\label{tab:robot_success_rate}
\resizebox{\columnwidth}{!}{
\begin{tabular}{lccc}
\toprule
Task & SFT Baseline & Qwen3-VL-$r_{prog}$ & LRM-$r_{prog}$ \\
\midrule
Giraffe-in-bowl & 38.3 (23/60) & 45.0 (27/60) & \textbf{56.7 (34/60)} \\
Can-on-cube & 43.3 (26/60) & 31.7 (19/60) & \textbf{55.0 (33/60)} \\
Open-drawer & 46.7 (28/60) & 18.3 (11/60) & \textbf{61.7 (37/60)} \\
Tissue-out-of-box  & 8.3 (5/60) & \textbf{95.0 (57/60)} & 88.3 (53/60) \\
\bottomrule
\end{tabular}
}
\end{table}

\section{CONCLUSIONS}

We introduced LRMs, which derive modality-specific contrastive, progress, and completion reward models from the same pretrained VLM. All three interfaces achieve higher mean success rates than the IL baseline in our simulation setting, with progress scoring performing best among the evaluated non-privileged rewards and narrowing the gap to privileged environment rewards. Across four real-world tasks on two robot platforms, LRM-scored BC consistently improves over SFT and provides more reliable gains across tasks than zero-shot Qwen3-VL scoring. These results support VLM reward specialization for online policy refinement and offline adaptation without hand-coded task rewards.






\section*{ACKNOWLEDGMENT}

The USC Physical Superintelligence Lab acknowledges generous supports from Toyota Research Institute, Dolby, Google DeepMind, Capital One, Nvidia, and Qualcomm. Yue Wang is also supported by a Powell Research Award.



\bibliographystyle{IEEEtran}
\bibliography{root}

\end{document}